\documentclass{article}

\usepackage{microtype}
\usepackage{graphicx}
\usepackage{subfigure}
\usepackage{booktabs} 

\usepackage{hyperref}

\usepackage[accepted]{icml2024}

\usepackage{amsmath}
\usepackage{amssymb}
\usepackage{mathtools}
\usepackage{amsthm}

\usepackage[capitalize,noabbrev]{cleveref}

\theoremstyle{plain}

\theoremstyle{definition}

\theoremstyle{remark}

\usepackage[textsize=tiny]{todonotes}
\newcommand{\innerproduct}[2]{\langle #1, #2 \rangle}

\crefname{equation}{}{}
\crefname{figure}{Figure}{Figures}
\creflabelformat{equation}{\textup{(#2#1#3)}}

\usepackage{enumitem}
\setlist[enumerate]{leftmargin=*,wide=0em, noitemsep,nolistsep, label = {\bfseries \arabic*.}}
\setlist[itemize]{leftmargin=*,wide=0em, noitemsep,nolistsep}

\icmltitlerunning{Manifold Integrated Gradients: Riemannian Geometry for Feature Attribution}

\begin{document}

\twocolumn[
\icmltitle{Manifold Integrated Gradients: Riemannian
Geometry for Feature Attribution}




\begin{icmlauthorlist}
\icmlauthor{Eslam Zaher}{CIRES,SMP}
\icmlauthor{Maciej Trzaskowski}{CIRES,IMB}
\icmlauthor{Quan Nguyen}{CIRES,IMB,QIMRB}
\icmlauthor{Fred Roosta}{CIRES,SMP}

\end{icmlauthorlist}

\icmlaffiliation{CIRES}{ARC Training Centre for Information Resilience (CIRES), Brisbane, Australia}
\icmlaffiliation{SMP}{School of Mathematics and Physics, University of Queensland, Brisbane, Australia.}
\icmlaffiliation{QIMRB}{QIMR Berghofer Medical Research Institute, Brisbane, Australia}
\icmlaffiliation{IMB}{Institute for Molecular Bioscience, University of Queensland, Brisbane, Australia}

\icmlcorrespondingauthor{Eslam Zaher}{e.zaher@uq.edu.au}

\icmlkeywords{Machine Learning, ICML}

\vskip 0.3in
]

\printAffiliationsAndNotice{}  

\begin{abstract}
In this paper, we dive into the reliability concerns of Integrated Gradients (IG), a prevalent feature attribution method for black-box deep learning models. We particularly address two predominant challenges associated with IG: the generation of noisy feature visualizations for vision models and the vulnerability to adversarial attributional attacks. Our approach involves an adaptation of path-based feature attribution, aligning the path of attribution more closely to the intrinsic geometry of the data manifold. Our experiments utilise deep generative models applied to several real-world image datasets. They demonstrate that IG along the geodesics conforms to the curved geometry of the Riemannian data manifold, generating more perceptually intuitive explanations and, subsequently, substantially increasing robustness to targeted attributional attacks.

\end{abstract}

\section{Introduction}
\label{submission}

As complexity of deep learning models continues to accelerate, ensuring their trustworthiness becomes paramount. Explainability has emerged as an important research field, making the behaviour of black-box deep learning more transparent. In this context,  feature attribution methods \citep{sundararajan2017axiomatic,StrivingSimplicityAll_2015,WhyShouldTrust_2016,UnifiedApproachInterpreting_2017,Saliency,PixelWiseExplanationsNonLinear_2015} represent a family of explainability techniques designed to attribute a model’s prediction to the most salient features in the input. These methods allow for the assessment of whether a model's decision is based on quantifiable and valid reasons, specifically in terms of the most influential factors that contribute to that decision.

Gradient-based feature attribution methods \citep{Saliency, StrivingSimplicityAll_2015, PixelWiseExplanationsNonLinear_2015} are commonly used, as they are more computationally efficient and faithful to the model’s internals, as opposed to other perturbation-based techniques \citep{WhyShouldTrust_2016, UnifiedApproachInterpreting_2017}. However, these methods suffer from two main drawbacks. First, the saliency maps produced for vision models often exhibit perceptual noise. Second, they are susceptible to adversarial attributional attacks.

Integrated gradients (IG) and other path-based methods are widely adopted because they satisfy axiomatic properties that are desirable in feature attribution methods. However, the choice of the path of attribution, and/or the baseline, impacts the quality and robustness of the generated explanations. For vision models, IG has faced specific criticism for accumulating noise along the integration path, diminishing the human-perceptual quality of the resulting explanations. A few studies attributed the main source of this noise to the ad hoc choice of the linear path of attribution. \citet{GuidedIntegratedGradients_2021} indicated that the linear path in IG is agnostic to the model output surface, leading to the accumulation of high-norm gradients assigned to irrelevant image pixels. \citet{InvestigatingSaturationEffects_2020} showed that gradients in saturated regions along the integration path, where the model output plateaus, can lead to disproportionate attributions. \citet{IDGIFrameworkEliminateExplanationNoise_2023} analyzed gradients at each point on the path, decomposing them into relevant and noise directions. Their results showed that including the noise direction in the path integral contributes considerably to the overall noise in the explanations.

Several methods have been proposed to mitigate the noise in the IG saliency maps. \citet{GuidedIntegratedGradients_2021} greedily optimized the path of attribution by guiding it through the model output surface. However, the resulting path from this method may stray into regions associated with adversarial examples due to its significant deviation from the straight-line path. Rather than utilizing the linear path in the image space, \citet{AttributionScaleSpace_2020a} integrated gradients along a continuum of blurred images. While this approach helps in reducing noise, it can also lead to a loss of fine-grained details due to the blurring effect. Split IG \citep{InvestigatingSaturationEffects_2020} excludes noisy saturated regions; however, it breaks the axioms fulfilled by IG. Other methods attempt to improve attributions by manipulating the input image and/or the baseline. SmoothGrad \citep{SmoothGradRemovingNoise_2017} averages attribution maps of noisy instances of the input image to generate less noisy feature visualisations. Alternative baselines \citep{sturmfels2020visualizing,
InterpretableExplanationsBlack_2017,SmoothGradRemovingNoise_2017} or even distributions of baselines \citep{lundstrom2022rigorous,ImprovingPerformanceDeep_2021}   have also been explored.

While the aforementioned methods aim at mitigating the noise issue in IG, there has been comparatively less emphasis on addressing IG's susceptibility to adversarial attributional attacks. \citet{InterpretationNeuralNetworks_2019} demonstrated that minimal perturbations to the input image can generate unstructured attributions due to the rapidly changing gradients at the decision boundary. IG has been shown to suffer from targeted attributional attacks, which can manipulate images to generate saliency maps that mimic explanations of arbitrary images, while maintaining a constant model output. Vulnerability to attributional attacks even apply to local model-agnostic methods \citep{FoolingLIMESHAP_2020} and global explanations \citep{FoolingNeuralNetwork_2019, FoolingPartialDependence_2023, FoolingSHAPStealthily_2022}. 
As a remedy to these attributional attacks, common approaches employ either one of two main strategies: (1) informing the classifiers about adversarial examples via adversarial training \cite{DeepLearningModels_2018, ConciseExplanationsNeural_2020}, (2) modifying the objective function to  maximise correlation of feature attributions between original and perturbed inputs \citep{ExplanationsCanBe_2019, RobustAttributionRegularization_2019}. There are also hybrid approaches that combine elements of both strategies \citep{FARGeneralFramework_2022, RobustAttributionRegularization_2019, ExploitingRelationshipKendall_2022}.

The bulk of work in adversarial robustness has substantiated that robust classifiers exhibit input-gradients that are semantically aligned with the human perception - a phenomena known as perceptually aligned gradients (PAGs) \citep{ArePerceptuallyAlignedGradients_2019, PerceptuallyAlignedGradients_2023, InputGradientsHighlight_2021, DeepLearningModels_2018}. Robustification using adversarial training is widely considered the most effective approach to rectify the model sensitivity to input perturbations, leading to both robust predictions and enhanced feature visualisations. The sharpness and robustness of feature visualizations from these classifiers is essentially a manifestation of PAGs. A few studies have shown that the data manifold is critical to achieving both robust classifiers and perceptual feature attributions. \citet{WhichModelsHave_2023} argued that for models to exhibit PAGs, they need to be more robust off the data manifold than on it. The resulting PAGs from off-manifold robustness are found to align closely with the data manifold. The interplay between the data manifold, PAGs, and gradient-based feature attribution methods has also been explored. \citet{ManifoldHypothesisGradientBased_2023} showed that the alignment of gradient-based saliency vectors with the data manifold corresponds to increased perceptual quality. Likewise, as models become more robust, their gradients tend to align more closely with the data manifold.

\textbf{Contributions.} Motivated by \citet{WhichModelsHave_2023, ManifoldHypothesisGradientBased_2023}, we take a data-manifold guided approach to simultaneously address both drawbacks of IG, i.e., the perceptual noise and the vulnerability to targeted attributional attacks. The main contributions of our work are as follows:
\begin{enumerate}[label = {\bfseries (\roman*)}]
    \item We introduce Manifold IG (MIG), a novel path-based attribution method that, by integrating along the geodesics of a latent Riemannian manifold, respects the curved geometry inherent to the underlying data manifold.
    \smallskip 
    \item Using deep generative models on real-world image datasets, we show that not only does MIG yield perceptually aligned feature visualizations but it also makes feature attributions more robust to targeted attributional attacks.
\end{enumerate}

\begin{figure*}[!htb]
\centering
\includegraphics[scale = 0.35]{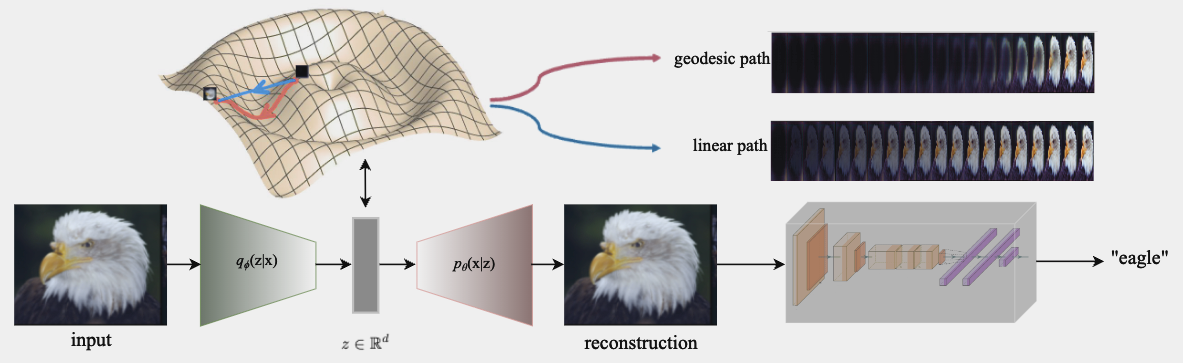}
\caption{Schematic of our Setup: The underlying image data manifold is learned using a convolutional VAE. The latent space corresponds to a Riemannian manifold where the geodesic path (shown in red) between two points represents the shortest path in such curved geometry. The linear path (shown in blue) doesn't conform to the intrinsic geometry of the manifold and deviates into regions out of the manifold. Reconstructions from the VAE along with the labels are used to train a classifier, and the geodesic path is used as the path of attribution in our MIG as opposed to the linear path in the image space used in IG.\label{fig:VAE_schematic}}
\end{figure*}

\section{Background}
We now briefly review several path-based feature attribution methods and provide some essential background necessary for our presentation.

\subsection{Path-based Feature Attribution Methods}
For target image $x$, baseline $x'$, and a classifier $F$, Gradient-based explainability methods are defined in terms of the derivative of $F$ with respect to the input. More specifically, \textit{path attribution methods} are a class of methods defined in terms of a path  $\gamma(t) := \gamma(x, x', t) $  connecting a baseline $x'$ and a target input $x$ by a continuous, smooth curve. In this light, path attribution methods are also baseline methods, and can be defined as 
\begin{align*}
\text{PathAttr}_j(x, x', \gamma) := \int_{0}^{1} \frac{\partial F(\gamma(t))}{\partial \gamma_j(t)} \frac{\partial \gamma_j(t)}{\partial t} \, dt.
\end{align*}

Depending on the choice of the baseline and the path of attribution, various approaches emerge. We now outline a few of these methods. 

\textbf{Integrated Gradients (IG).} the original path-based feature attribution method \citep{sundararajan2017axiomatic} with the simplest linear path function $\gamma(t) = x' + t(x -x')$ as
\begin{align*}
\text{IG}_j(x, x') := (x_j - x'_j) \times \int_{t=0}^{1} \frac{\partial F(x' + t(x-x'))}{\partial x_j} dt.
\end{align*}

The baseline is commonly chosen to be either a black or white image in the case of vision models. IG aggregates contributions of features along a path from a baseline with an absolute absence of influential features, and progressively increases the presence of the feature signal along the path until reaching full abundance at the target feature value.

\textbf{Guided Integrated Gradients (GIG).} \citet{GuidedIntegratedGradients_2021} showed that IG generates noisy attributions due to accumulation of noise along the linear path. They generalized the path function to be guided by the model output surface, allowing the path of attribution to avoid regions of saturated and high gradients. GIG can be expressed as
\begin{align*}
\text{GIG}_j(x, x') := \int_{t=0}^{1} \frac{\partial F(\gamma^F (t))}{\partial \gamma_j ^F (t)} \frac{\partial \gamma_j ^F (t)}{\partial t} \, dt,
\end{align*}
where $\gamma^F (t) := \gamma(x, x', F, t)$ represents the path guided by the model, which is greedily optimized to minimize the impact of high-norm model gradients.

\textbf{Blur IG.} This method can be viewed as a variation of IG on a path that transitions from a fully blurred image to the original image \citep{AttributionScaleSpace_2020a}, and is given by 
\begin{align*}
\text{BlurIG}(m, n) = \int_{t=\infty}^{0} \frac{\partial F(L(m, n, t))}{\partial L(m, n, t)} \frac{\partial L(m, n, t)}{\partial t} dt,
\end{align*}

where $m,n$ are the pixel indices in the image, and $L$ is a 2D Gaussian blur kernel with variance $t$ as 
\begin{align*}
    L(m, n, t) = \sum\limits_{k=-\infty}^{\infty} \sum\limits_{l=-\infty}^{\infty} \frac{1}{\pi t} e^{-\frac{m^2 + n^2}{t^2}} x(m - k, n - l).
\end{align*}

\subsection{Attributional Attacks}

The susceptibility of feature attribution methods to adversarial attributional attacks raises concerns about their reliability. \citet{InterpretationNeuralNetworks_2019} demonstrated that adding systematic human-imperceptible perturbations to the input image can cause amorphous change in the feature maps, while maintaining the same output class. \citet{ExplanationsCanBe_2019}  proposed a targeted attack to generate feature explanations that can match any arbitrary target feature map. They showed that the model output surface in ReLU-based image classifiers exhibits high curvature, causing increased vulnerability of the gradient-based explanations. To enhance robustness of explanations, they smoothed the model's output curvature by replacing the ReLU activations in the model with SoftPlus at the time of generating attributions. 

Among the common attacks in this context are top-k and targeted attributional attacks.

\textbf{Top-k Attributional Attack.} This approach generates an explanation that reduces the attribution score of the highest $k$ features in the original feature map. \citet{InterpretationNeuralNetworks_2019} defines the top-k attack as $x_{\text{adv}}^* = \arg\min_{x_{\text{adv}}} C(I(x), I(x_{\text{adv}}))$ subject to the constraints that $\|x_{\text{adv}} - x\|_{\infty} \leq \epsilon$ and $F(x_{\text{adv}}) = F(x)$, 
where $I$ is the attribution method of interest, $C(x, x_{\text{adv}}) = \sum_{j \in K} I_j(x)$ , $K$ is the set of top $k$ features in $x_{\text{adv}}$, and $\|.\|_{\infty}$ denotes the infinity norm on vectors.

\textbf{Targeted Attributional Attack.} This method manipulates the input image to generate attributions that resemble feature maps corresponding to a different target image \citep{ExplanationsCanBe_2019}. It can be formulated as $x_{\text{adv}}^* =  \arg \min_{x_{\text{adv}}} \mathcal{L}(x_{\text{adv}})$ 
where $\mathcal{L}(x_{\text{adv}}) = \left\| I(x_{\text{adv}}) - I(x_{\text{target}}) \right\|_{2}^2 + \gamma \left\|F(x_{\text{adv}}) - F(x) \right\|_{2}^2$, 
with $\gamma$ and $x_{\text{target}}$ being, respectively, a hyperparameter that controls the relative weight of the two summands and the target image. Here, $\|.\|_{2}$ signifies the Euclidean norm.

\subsection{VAEs for Generative Manifold Learning}
A substantial body of literature on manifold learning relies on neural autoencoders (AEs) \citep{AutoencodersMinimumDescription_1993, ExtractingComposingRobust_2008}. The underlying hypothesis guiding the success of manifold learning is that even though the data may exist in a high-dimensional space, it effectively lies on or near an embedded low-dimensional manifold \citep{TestingManifoldHypothesis_2016, RepresentationLearningReview_2013, WhyDeepLearning_2016}. 

Unlike most traditional AEs that lack structure over the latent space, variational autoencoders (VAEs) are the probabilistic variants with a regularised, generative latent structure \citep{AutoEncodingVariationalBayes_2022, BetaVAELearningBasic_2016, UnderstandingDisentanglingBeta_2018}. VAEs aim not only to learn a compressed representation but also to model the underlying probability distribution of the data in a lower-dimensional latent space. This latent space is typically assumed to adhere to a prior distribution, often modeled as a multivariate Gaussian. This allows VAEs to capture the intrinsic geometrical and statistical structure of the data using more informative latent representations. The learned latent space is then used to generate new data points similar to the training data, making VAEs suitable for tasks such as image generation and reconstruction, interpolation, and out of distribution detection \citep{DetectingOutofdistributionSamples_2022, GeneratingInBetweenImages_2020}. 

High-dimensional data such as real images are assumed to have a non-Euclidean latent structure that constitutes the natural image manifold \citep{GenerativeVisualManipulation_2016}. Nevertheless, the data is often perceived through the lens of Euclidean geometry as it offers definite inner products and explicit distance metrics, making manipulation of data more accessible. \citet{RiemannianGeometryDeep_2018} showed that nonlinear VAEs induce a Riemannian metric over the latent space. Similarly, \citet{LatentSpaceOddity_2018} demonstrated that the induced Riemannian metric leads to faithful latent-space statistical estimates, smooth interpolations, and better generalisations. This motivated a compilation of methods to learn non-Euclidean latent structures, e.g.,   hyperspherical \citep{HypersphericalVariationalAutoEncoders_2018}, hyperbolic \citep{PoincarGloVeHyperbolic_2018, ContinuousHierarchicalRepresentations_2019, HyperbolicVAELatent_2023}, mixed-curvature \citep{MixedcurvatureVariationalAutoencoders_2019} and Riemannian weighted submanifolds \citep{LearningWeightedSubmanifolds_2020}.

\section{Riemannian Geometry for Feature Attribution}
We now present our proposed feature attribution method. To that end, we present elements of differential geometry as they apply to deep generative models. We demonstrate how VAEs induce a Riemannian metric over the latent space. Utilizing the induced metric, we use an algorithm to compute geodesic paths that conform to the curved structure of the data manifold. Following this, we introduce Manifold Integrated Gradients (MIG) for feature attribution by integrating model gradients along these geodesics.

\subsection{Latent Space Geometry in Deep Generative Models}
A smooth manifold $\mathcal{M}$  is a topological manifold with a smooth structure that is locally homeomorphic to Euclidean space \citep{lee2012smooth}. This implies that around any point $z \in \mathcal{M}$ on the manifold, $\mathcal{M}$ resembles $\mathbb{R}^{d}$, enabling to extend differential calculus on the manifold. On smooth manifolds, concepts like length, inner products, and shortest paths, known as geodesics,  take on a more complex nature compared to their Euclidean counterparts. 

The governing mathematical apparatus that generalizes these concept to manifolds and gives a formal framework to compute them is the \textit{Riemannian metric} \citep{lee2018introduction}. Recall that a tangent space of $\mathcal{M}$ at $z$, denoted by $T_z \mathcal{M}$, is a vector space spanning all the tangent vectors to $z$ on $\mathcal{M}$. Since $T_z \mathcal{M}$ is a vector space, we can define an inner products on it. If this inner product varies smoothly with $z$, then it defines a Riemannian metric. More specifically, a Riemannian metric on a smooth manifold $\mathcal{M}$ assigns to each point $z \in \mathcal{M}$ an inner product $\innerproduct{\ }{\ }_z: T_z \mathcal{M} \times T_z \mathcal{M} \to \mathbb{R} $ that varies smoothly on the manifold.
A \textit{Riemannian manifold} is a smooth manifold with a Riemannian metric \citep{lee2018introduction}.

Latent-variable generative models with smooth generator functions embody surface models \citep{LatentSpaceOddity_2018}. They incorporate a generative function: $g: \mathcal{M} \to \mathcal{X} $, which transforms a latent manifold $\mathcal{M} $ in the latent space $\mathcal{Z} \subset \mathbb{R}^{d}$ into a data manifold embedded in the data space  $\mathcal{X} \subset \mathbb{R}^{D}$. In the VAE setting, the decoder acts as the generator function, and  the latent space is chosen to have a lower dimensionality  compared to the input space, i.e., $d \leq D$. 

A smooth latent curve connecting points $z_{0} \in \mathcal{M} $ and $z_{1} \in \mathcal{M} $ can be parameterized by a function $\gamma: [0, 1] \to \mathcal{M}$ with $\gamma(0) = z_0$ and $\gamma(1) = z_1$. This is then mapped by the generative function to a corresponding smooth curve $g (\gamma): [0, 1] \to \mathcal{X}$ on the data manifold. The length of the curve $g(\gamma)$ can be expressed as
\begin{align*}
    L(g (\gamma)) &= \int_0^1 \left\| \frac{d g (\gamma(t))}{d t} \right\| dt \\
    &= \int_0^1 \left\| J_g(\gamma(t))\gamma^{\prime}(t) \right\| dt,
\end{align*}
where $J_g(\gamma(t)) = {\partial g(z)}/{\partial z} \mid_{z= \gamma(t)}$ is the Jacobian of $g$ at $\gamma(t)$, and $\gamma^{\prime}(t) = {\partial\gamma(t)}/{\partial t}$ is the velocity vector tangential to the latent curve.
This can in turn be written as
\begin{align*}
L(g (\gamma)) &= \int_0^1 \sqrt{(J_g (\gamma(t)) \cdot \gamma'(t))^{T} \cdot (J_g (\gamma(t)) \cdot \gamma'(t))} \, dt \nonumber \\
&= \int_0^1 \sqrt{\gamma'(t)^T G_g(\gamma(t)) \gamma'(t)} \, dt,
\end{align*}
where $G_g(.) = J_g^T(.) J_g(.) $. Under certain architectural choices \citep{RiemannianGeometryDeep_2018}, $G_g(.)$ is symmetric positive definite and smooth on $\mathcal{M}$, and hence defines a Riemannian metric, which can in turn be used to calculate geodesics, i.e., the shortest path between two points on $\mathcal{M}$.

\begin{figure}[htbp]
  \centering
  \includegraphics[width=\linewidth]{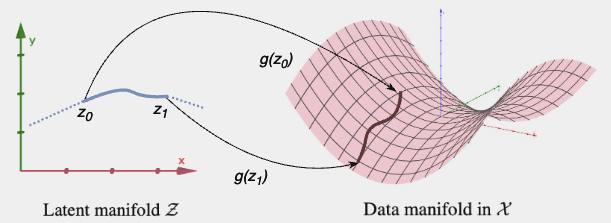}
  \caption{The surface model implied by the smooth generator function $g$ mapping from the latent space $\mathcal{Z}$ to the data space $\mathcal{X}$. In this example, the latent manifold $\mathcal{M}$ is a one-dimensional embedded submanifold of $\mathbb{R}^{2}$ and the images lie on a two-dimensional embedded submanifold of $\mathbb{R}^{3}$. The geodesic on the latent manifold is mapped to a smooth curve on the data manifold, respecting the underlying geometry.}
  \label{fig:surface_model}
\end{figure}

The  length functional, $L(g (\gamma))$, is invariant under reparameterization \citep[Lemma 1.4.3]{jost2008riemannian}. Hence, to find curves of shortest length on $\mathcal{M}$, we can simply consider curves that are parameterized proportionally by arc length. Consequently, it can be shown that the following energy functional
\begin{align}
\label{eq:energy}
E(g(\gamma)) = \frac{1}{2} \int_0^1 \gamma^{\prime}(t)^T G_g(\gamma(t)) \gamma^{\prime}(t) \, dt,
\end{align}
is essentially equivalent to $L^{2}(g (\gamma))$. Indeed, while in general, $L^{2}(g (\gamma))\leq 2E(g(\gamma))$, the equality holds if and only if $\|d g(\gamma(t))/dt\|$ is constant \citep[Lemma 1.4.2]{jost2008riemannian}. This, gives rise to the following minimization problem 
\begin{subequations}
\label{eq:energy_min}
\begin{align}
\min_{\gamma \in \Gamma} & \; E(g (\gamma)), 
\end{align}
where
\begin{align}
    \Gamma = \Big\{\gamma: [0,1] \to \mathcal{M} \mid \gamma(0) = z_0, \; \gamma(1) = z_1, \; \text{ and }\Big. \nonumber\\
    \Big.\left\|{d g(\gamma(t))}/{dt}\right\| \equiv \text{const}\Big\},
\end{align}
\end{subequations}
and whose solution $\gamma^{*}$ is the shortest path, or geodesic, between $z_0$ and $z_1$ on $\mathcal{M}$. In this light, traversing along the shortest path from $z_0$ to $z_1$ on the latent manifold amounts to a smooth transition from $g(z_0)$ to $g(z_1)$ on the data manifold in the sense of \cref{eq:energy_min}; see \cref{fig:surface_model}. 

It can be shown that the curve $\gamma^{*} (t)$ is a solution to \cref{eq:energy_min} if and only if it satisfies the system of second-order ordinary differential equations \citep[Lemma 1.4.4]{jost2008riemannian}
\begin{align}
\label{eq:eu-lg}
\frac{d^2 \gamma^k(t)}{dt^2} + \sum_{i,j} \Gamma^k_{ij} \frac{d\gamma^i(t)}{dt} \frac{d\gamma^j(t)}{dt} = 0,
\end{align}
where $k = 1, \ldots, d$, and $\Gamma^{k}_{ij}$ are the Christoffel symbols associated to the Riemannian metric tensor $G_g = (g_{ij})_{i,j = 1,\ldots,d}$. The Christoffel symbols are  defined as
\begin{align}
\label{eq:chris-symbs}
\Gamma^{k}_{ij} = \frac{1}{2} \sum_{\ell=1}^{d}  g^{k\ell}\left( \frac{\partial g_{j \ell}}{\partial x^{i}} + \frac{\partial g_{i\ell}}{\partial x^{j}} - \frac{\partial g_{ij}}{\partial x^{\ell}} \right),
\end{align}
where $(g^{ij})=(g_{ij})^{-1}$, i.e., $\sum_{\ell=1}^{d} g^{i\ell}g_{\ell j} = \delta_{ij}$.

Previous works have proposed different approaches to calculate geodesic paths for generative models. \citet{RiemannianGeometryDeep_2018} used a discretization of $\gamma$ and a finite-difference formulation of the energy functional (\ref{eq:energy}) to avoid the computational burden of the Euler-Lagrange system of equations (\ref{eq:eu-lg}). \citet{LatentSpaceOddity_2018} computed the Christoffel symbols (\ref{eq:chris-symbs}) and then solved the second order system numerically to get geodesic paths for VAEs with stochastic generators. \citet{FastRobustShortest_2019} employed a scheme based on fixed-point iterations to solve the system in (\ref{eq:eu-lg}) without computing Jacobians that are usually ill-conditioned. In this work, we use a slight modification of \citet[Algorithm 1]{RiemannianGeometryDeep_2018} to calculate geodesic paths; see \cref{sec:appendix_a} for details.

\subsection{Integrated Gradients on the Data Manifold}
Our generative-discriminative approach is depicted in \cref{fig:VAE_schematic}. We train a VAE to capture the underlying Riemannian data manifold for real image datasets. We feed-forward the whole datasets through the VAE and use the resulting reconstructions from the learned image data manifold along with the ground-truth labels to train a deep convolutional classifier. We then exploit the intrinsic geometry of the manifold to build a path-based feature attribution method by integrating gradients of the classifier’s output along geodesics on the manifold.
This leads to defining MIG along the geodesic path, $\gamma^*$ from \cref{eq:energy_min}, for the $j^{th}$ feature as
\begin{align*}
\text{MIG}_j(x, \gamma^{*}) := \int_{0}^{1} \frac{\partial F(g(\gamma^{*}(t)))}{\partial g_j(\gamma^{*}(t))} \frac{\partial g_j(\gamma^{*}(t))}{\partial t} \, dt.
\end{align*}

MIG employs the smoothest path between a baseline and a target in the sense of \cref{eq:energy_min}. In essence, the smooth generator maps the geodesic on the latent space to a highly smooth path on the data manifold that respects the underlying curved geometry. Further, like any path-based attribution method, MIG satisfies the axioms of completeness, sensitivity, and implementation invariance \citep{sundararajan2017axiomatic}.

\section{Experiments}
\label{sec:exp}
We now evaluate MIG in several contexts. In \cref{sec:geo}, we first demonstrate the effect of smooth interpolating paths adherent to the underlying geometry that underpins MIG. In \cref{sec:perc}, we compare various methods with MIG in terms of the perceptual alignment of their feature attribution maps.  Finally, in \cref{sec:adv}, we investigate MIG in the context of targeted attributional attacks. 
The code for these experiments is available \href{https://github.com/eszaher/Manifold-Integrated-Gradients}{here}.
\begin{figure}[!htb]
  \centering
  \includegraphics[width=\linewidth]{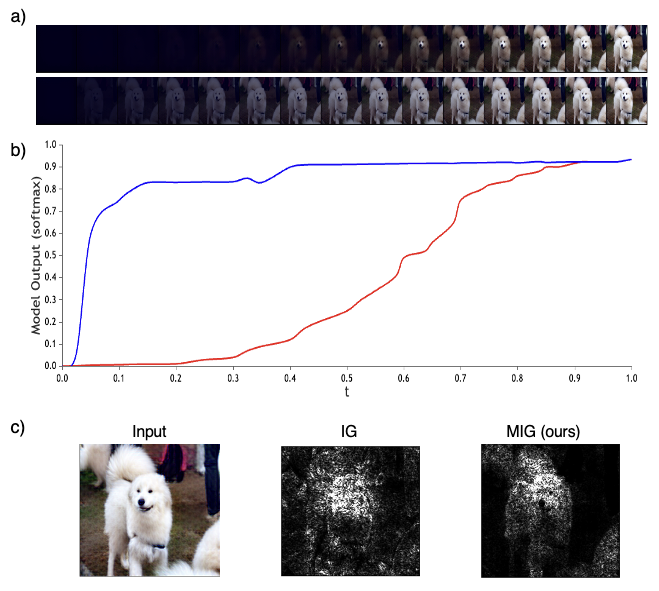}
  \caption{Mapped geodesic interpolation in MIG vs.\ linear interpolation in IG. (a) contrasts the smoothness of MIG's smooth path interpolants against IG's linear path from a black baseline. (b) displays a classifier response curves for each image on the paths, with MIG's smooth path (red) having a gradual response as key features show later on the path and IG's linear path (blue) showing rapid escalation, with a wide saturation region. (c) shows the corresponding feature visualizations. MIG produces more perceptually aligned and less noisy feature visualizations compared to IG.}
  \label{fig:interpolation}
\end{figure}
\begin{figure*}[!htb]
\centering
\includegraphics[width=\textwidth]{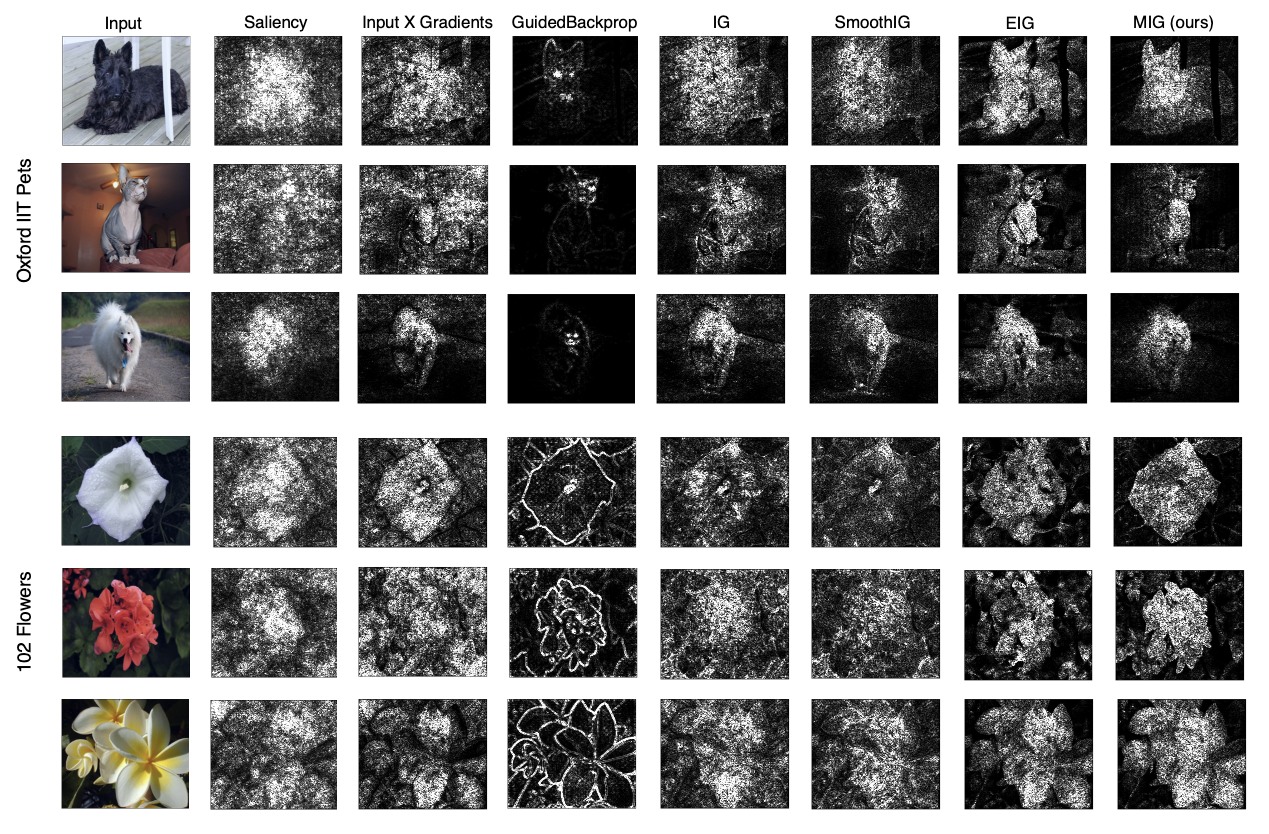}
\caption{ Comparison of Feature Attribution Methods. Presented are feature maps from various methods—Saliency, Gradients $\times$ Input, GuidedBackprop, IG, Smooth IG, EIG, as well as our proposed MIG. As shown, MIG addresses the IG's noise limitation and surpasses other methods, producing distinctly clearer and perceptually more aligned visualizations. In the last row, the similarity between EIG and MIG indicates that the path of attribution in MIG passes through a nearly flat region on the data manifold, and hence a linear interpolation path employed by EIG can closely approximate the mapped geodesic path in MIG for this particular image.}
\label{fig:attributions}
\end{figure*}
\begin{figure*}[!htb]
\centering
\includegraphics[width=\textwidth]{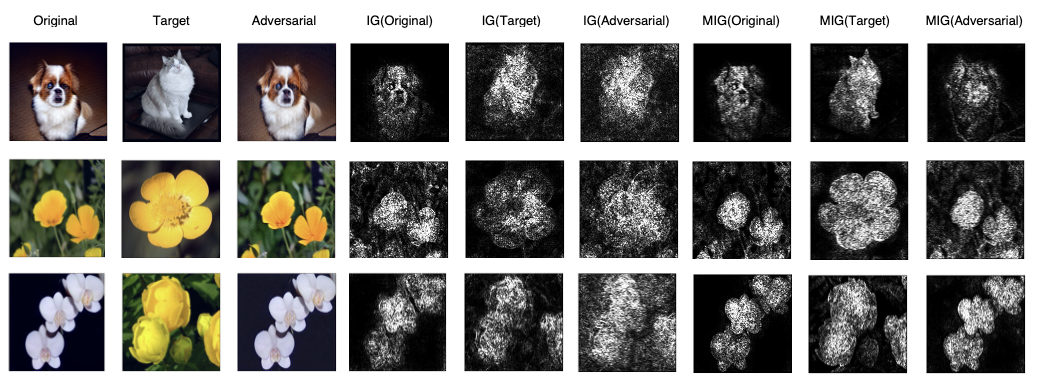}
\caption{ MIG vs.\ IG under targeted attributional attacks. The figure displays examples of an original input image alongside a target image and an adversarial attributional attack designed to exploit the IG's linear path of attribution.  IG's vulnerability is evident as it generates adversarial feature maps that erroneously mimic the target maps. MIG maintains perceptually consistent and noise-resistant feature visualizations for the adversarial examples, closely resembling those of the original input. Each row was generated based on a different classifier's backbone, VGG-16, ResNet18, InceptionV1, respectively.}
\label{fig:robustness}
\end{figure*}

\textbf{Setup.} 
For learning the image data manifold, we employ a convolutional VAE based on residual connections; see \cref{sec:appendix_b} for details. We extend the typical Evidence Lower Bound (ELBO) loss by adding a perceptual loss term \citep{DeepFeatureConsistent_2016} that emphasises feature-wise accuracy. Training VAEs with perceptual loss helps to mitigate the inherent blurriness in reconstructions as it preserves structural and perceptual details. We train the VAEs for 150 epochs with Adam \citep{kingma2014adam}. We then use the image reconstructions along with the labels to train classifiers based on pretrained models.  

For showing the perceptual quality of feature visualizations and robustness to targeted attributional attacks, we utilize different backbones (VGG-16, Resnet-18, InceptionV1) for the discriminative models. All the backbones are frozen during training the classifiers for the first 10 epochs, before fine-tuning the whole models for another 7 epochs.

\textbf{Datasets.}
We validate our approach on two real-image datasets: (1) the Oxford-IIIT Pet Dataset \citep{oxford-iiit-pet}, which compromises pet images of 37 categories, making it well-suited for fine-grained classification tasks. The images vary in pose and lighting, making it also apt for the task of examining how VAEs capture and represent the inherent properties of the underlying data manifold. (2) the Oxford 102 Flower Dataset \citep{oxford-102-flowers}, which is also used for fine-grained recognition tasks. It compromises 102 different flower categories, with a large variation in scale, pose, color, and background, reflecting the intricacies in real-world settings. This motivates our use of this dataset, as we believe that VAEs can effectively capture a data manifold with such complexity and intra-class variability.\

To ensure a fair comparison of our method with IG, we need to use the standard black and white baseline images. Our goal is to generate geodesic paths of attribution on the manifold, linking these baseline images to the target images. However, due to the variational approximation in VAEs and their inherent noise, reconstructions of purely black or white images is not feasible. To incentivise the VAEs to produce cleaner black and white baseline reconstructions, we augment all datasets with black and white images during the training phase of the VAEs. This ensures VAEs can handle such plain images.

\textbf{Baseline Methods.} We compare our method against the following alternatives: 
\begin{itemize}
    \item \textbf{Saliency} \citep{Saliency}. The simplest method for capturing sensitivity of the model to changes in the input. 
    \item \textbf{Input $\times$ Gradients} \citep{LearningImportantFeatures_2017}. A method proposed to add sharpness to the sensitivity maps. 
    \item \textbf{Guided Backpropagation} \citep{StrivingSimplicityAll_2015}. It uses a modification to gradient calculations in ReLU-based classifiers by only backpropagating non-negative gradients.
    \item \textbf{IG} \citep{sundararajan2017axiomatic}. The original path-based method that uses the linear path in the image space.
    \item \textbf{Smooth IG} \citep{SmoothGradRemovingNoise_2017}: It reduces the noise in IG by averaging feature attributions from noisy images.
    \item \textbf{EIG} \citep{EnhancedIntegratedGradients_2020}. In this approach, the manifold is conceptualized as a flat, Euclidean space, and it generates feature maps along straight linear paths in the latent space.
\end{itemize}

\subsection{Geodesic Paths of Attribution}
\label{sec:geo}
In curved manifolds, geodesics are the shortest paths that represent the most seamless transitions possible between two points. When applying this to latent representation of images, it translates to a smooth interpolation between two images, staying faithful to the structure of the data manifold. This means that the transformed geodesic path yields a series of realistic images with minimal perceptible differences between successive interpolations. In contrast, a linear interpolation path, based on Euclidean geometry, tends to deviate outside the data manifold, violating the curved nature of the underlying space. 

The linear path in IG, $\gamma = x' + t(x-x')$, transforms all pixels with the same scale, which implies a complete  independence amongst pixels, violating real-world settings.
The progression of pixels along a smooth path on the data manifold, however, is a complex and interdependent process. This intricacy stems from the interconnected nature of pixels, where each pixel does not exist in isolation. The essence lies in the idea that ``\textit{the whole is greater than the sum of its parts},'' as the transitions are shaped by the influence of neighboring pixels. In \cref{fig:interpolation}(a), we can see that superpixels on the smooth path surrounding dog's nose start to arise first, and pixels affect relevant pixels as they progress, until we reach the full target image. 
It is also worth highlighting that specific feature structure (the nose in this case) can emerge and evolve differently from other features, which means that progression is curved and context-dependent. This curvature in the data manifold is influenced by the distribution of intensities, colors, gradients within the image, and structural properties in the image, making the mapped geodesic path a complex function of all those influential elements and variations. In this sense, MIG along smooth paths on the manifold can capture feature interactions, which is a major limitation of the linear path in IG.

From the perspective of the model output, shown in \cref{fig:interpolation}(b), mapped geodesics avoid the wide saturation region, where the model's response becomes less sensitive to changes in input features. As gradients saturate in these regions, the resulting attributions are predominantly influenced by noise. The linear path in IG is a significant contributing factor to the issue of saturation-induced noise \citep{GuidedIntegratedGradients_2021, InvestigatingSaturationEffects_2020}. This arises because the linear path inherently crosses through a wide saturation region. The essence of the problem lies in the nature of linear interpolation, which introduces discriminative image features early in the path. As a result, the model's response escalates rapidly at the beginning of the linear path and then quickly plateaus. The corresponding feature maps to the two paths of attribution are shown in \cref{fig:interpolation}(c).

\subsection{Perceptual Attribution Maps along the Geodesics}
\label{sec:perc}
The underlying data manifold provides the natural structure for the model to learn. When model gradients are aligned with this image manifold, it ensures that gradients are perceptually relevant. Geodesics come into play to enforce integrated gradients to be aligned with the manifold, emphasising features that are consistent with human perception. MIG, utilising transformed geodesic paths that conform to the curved natural structure of the data manifold, is essentially an accumulation of PAGs. In \cref{fig:attributions}, we show feature attribution maps generated by MIG (last column) as compared to other methods. In all examples, MIG clearly outperforms others as it produces perceptually consistent attributions with minimal noise. 
In the last row in \cref{fig:attributions}, one can see a certain degree of similarity between EIG and MIG feature maps. This is because, for this specific image, the linear path employed by EIG is an approximation to the geodesic due to the low curvature of the latent manifold along the geodesic from the baseline to the target.

\subsection{Robustness to Targeted Attributional Attacks}
\label{sec:adv}
Adversarial attributional attacks exploit model's sensitivity to imperceptible perturbations, which are usually off-manifold. Aligning the model's gradients with the data manifold, provides inherent robustness to such perturbations. The linear path of attribution in IG adds up to this vulnerability as it results in interpolants on the path being positioned outside the manifold. This shift of interpolants away from the manifold can potentially move them into regions of adversarial examples, leading to the accumulation of irregular gradients. VAEs possess denoising capabilities, allowing them to denoise input, including noisy or adversarial examples. This process can be seen as projecting noisy or adversarial inputs closer to the manifold of normal data. MIG, by adhering to the data manifold, increases robustness. \cref{fig:robustness} shows how MIG feature maps are more robust to targeted attributional attacks as compared to IG. In this case, an input image is manipulated to compose an adversarial example. This generates attributions similar to an arbitrary target image, while maintaining the same output class as the original input. Unlike employing adversarial training to build robust classifiers that exhibit PAGs – a process that is computationally intensive and often detrimental to the model's performance \citep{RobustnessMayBe_2018} – our approach focuses on learning the data manifold and aligning model gradients with the intrinsic geometry. This approach effectively generates robust and perceptually aligned feature visualizations, achieving this without sacrificing the accuracy of the classifier, though it introduces some additional complexity as highlighted in \cref{sec:appendix_c}. 
\section{Quantitative Analysis}
\label{sec:quantitative}

To assess our method (MIG), we employ robust metrics from \citet{FidelitySensitivityExplanations_2019} and \citet{1284395} to evaluate the generated explanations. We compare MIG against IG variants, examining fidelity, sensitivity, and robustness through metrics such as infidelity, maximum sensitivity, and structural similarity to validate our method’s effectiveness.

\subsection{Metrics}
\paragraph{Explanation Infidelity (INFD) \citep{FidelitySensitivityExplanations_2019}.}
This measure is a robust variant of the \textit{completeness} axiom. It quantifies the degree to which an explanation misrepresents the model's sensitivity to input perturbations, assessing inaccuracies reflected in feature attributions. This contrasts with the completeness axiom that requires feature attributions to sum up to the total change in model output, without directly evaluating perturbation sensitivity. For a black-box function $f$, an explanation functional $\Phi$, and a random vector $\xi$ that characterizes significant perturbations of interest around input $x$, the explanation infidelity of $\Phi$ is defined as
\begin{align*}
\text{INFD}(f,x,\Phi) = \mathbb{E}_{\xi} \left[  \langle \xi, \Phi(f, x)\rangle - (f(x) - f(x - \xi)) \right]^{2},
\end{align*}
where  $\langle.,.\rangle$ denotes the Euclidean inner-product between vectors. One plausible way to determine $\xi$ is by calculating the difference from a noisy baseline: \(\xi = x - z_0\), where \(z_0 = x_0 + \epsilon\). In this context, \(\epsilon\) denotes a zero-mean random vector, such as \(\epsilon \sim \mathcal{N}(0, \sigma^2 I )\).

\paragraph{Maximum Sensitivity ($\textbf{SENS}_{\textbf{max}}$) \citep{FidelitySensitivityExplanations_2019}.}
The essence of max-sensitivity is to evaluate how sensitive an explanation is to perturbations in the input, with a particular focus on its maximum deviation. Given an input neighborhood radius r, maximum sensitivity is defined as
\begin{align*}
\text{SENS}_{\text{max}} = \max_{\|\delta\|\leq r} & \left\| \Phi(f(x + \delta)) - \Phi(f(x)) \right\|,
\end{align*}
where $\|.\|$ is usual Euclidean norm of the vectorized input.  It is worth noting that while reducing sensitivity might seem desirable to enhance robustness, doing so without careful consideration can lead to sub-optimal feature attributions. For instance, the intrinsic sensitivity of natural explanations—either due to the model's inherent sensitivity or the explainability method—suggests that some degree of sensitivity is unavoidable and, in fact, necessary to maintain the fidelity of the explanation to the model's behavior. However, it is possible to reduce the sensitivity of an explanation in a way that also lowers its infidelity. This dual benefit is significant, as it suggests that reliable explanations can be less sensitive, and also more accurate in representing the model's predictive behavior.

\paragraph{Structural Similarity Index (SSI) \citep{1284395}.} We use this metric to assess MIG's robustness to IG-targeted attributional attacks. SSI can measure the similarity between attribution maps for the input and its adversarial version as it captures intricacies be evaluating the match in contrast levels, brightness (attribution scores), and structures within the attribution maps.

\subsection{Results}

Our method satisfies the axioms of path-based feature attribution while also ensuring reliability and faithfulness with robust metrics. \cref{table:INFD} indicates that MIG achieves the lowest maximum sensitivity to input perturbations and significantly enhances faithfulness compared to other methods. This is viewed through the data-manifold perspective, where perturbations typically displace the input from the manifold. Our geodesic path of attribution, aligning with the manifold’s geometry, renders the feature attribution maps less sensitive to perturbations and more faithful to the model’s output behavior.

\begin{table}[h]
\centering
\begin{tabular}{|l|c|c|c|c|}
\hline
Datasets & \multicolumn{2}{c|}{Oxford IIT Pets} & \multicolumn{2}{c|}{102 Flowers} \\
\hline
Methods & SENS\textsubscript{max} & INFD & SENS\textsubscript{max} & INFD \\
\hline
IG    & 0.87  & 7.65 & 0.74  & 15.26 \\
BlurIG    & 0.75  & 6.41 & 0.67  & 12.19 \\
SmoothIG & 0.42  & 4.30 & 0.38  & 9.81 \\
MIG(ours) & \textbf{0.17}  & \textbf{1.86} & \textbf{0.21}  & \textbf{3.46} \\
\hline
\end{tabular}
\caption{Sensitivity and Infidelity of Feature Attributions. Here, lower values signify better quality. Our method, MIG, achieves the highest quality among the alternatives.}
\label{table:INFD}
\end{table}

\begin{table}[h]
\centering
\setlength{\tabcolsep}{5pt} 

\begin{tabular}{|l|c|c|c|c|c|c|}
\hline
\multicolumn{1}{|l|}{\text{Datasets}} & \multicolumn{3}{c|}{\text{Oxford IIT Pets}} & \multicolumn{3}{c|}{\text{102 Flowers}} \\ \hline
\text{Backbones} & \multicolumn{3}{c|}{SSI} & \multicolumn{3}{c|}{SSI} \\ \cline{2-7}
        & IG     & SIG    & MIG    & IG     & SIG    & MIG     \\ \hline
VGG-16   &   0.43   &   0.64   &   \textbf{0.87}   &    0.35   &    0.57  &   \textbf{0.74}   \\
ResNet18 &   0.48   &   0.69   &   \textbf{0.91}   &    0.41   &    0.63  &   \textbf{0.86}   \\
InceptionV1 &  0.36   & 0.61   &   \textbf{0.86}   &     0.36  &    0.52  &   \textbf{0.71}   \\ \hline
\end{tabular}
\caption{Structural Similarity Under IG-targeted Attributional Attacks. Here, higher values signify more robustness. Our method, MIG, achieves the greatest robustness among the alternatives.}
\label{table:SSI}
\end{table}

\noindent

\cref{table:SSI} shows that MIG scores the highest SSI, indicating it preserves the structure and relevancy of attribution maps under targeted attributional attacks. Unlike IG, which uses a noninformative linear path accumulating adversarial gradients, and SIG, which averages multiple linear paths potentially smoothing feature maps for added robustness, MIG avoids these linear paths. Instead, it uses a single geodesic path that exhibits a higher degree of robustness.

\section{Conclusion}

Our work introduces MIG as a solution to the reliability issues in IG for deep learning models. MIG leverages geodesics on the latent manifold to provide smoother interpolations between images, capturing the non-linear nature of image manifolds. In contrast to the linear path in IG, MIG captures pixel interactions more realistically, reducing noise in feature attributions. Additionally, MIG enhances model's robustness against targeted attributional attacks by aligning gradients with the data manifold. Our experiments validate the effectiveness of MIG, offering perceptually aligned explanations and promising ``safer'' applications in domains requiring enhanced model interpretability and reliability, e.g., critical sectors such as healthcare.

\section*{Acknowledgements}
This research was partially supported by the Australian Research Council through an Industrial Transformation Training Centre for Information Resilience (IC200100022). Quan Nguyen is supported by a NHMRC Investigator Grant (GNT2008928).
\section*{Impact Statement}

This paper presents MIG, an approach that can potentially be applied to improve interpretability in a broad range of high-risk neural network applications. It complements the existing discussions on Explainable AI from a data-manifold perspective, and bridges the gap between reliability and effectiveness of gradient-based explainability techniques. A direct positive impact of the proposed method is to enhance transparency and accelerate adoption of complex vision models for medical imaging. However, while we enhance robustness to targeted attributional attacks, potentially new adversarial attacks can target our method and exploit vulnerabilities that we did not address.

\bibliography{main}
\bibliographystyle{icml2024}

\newpage
\appendix
\onecolumn
\section{Further Details on Computing Geodesics}
\label{sec:appendix_a}

For the sake of self-containment, we briefly review the approach taken by \citet[Algorithm 1]{RiemannianGeometryDeep_2018} to compute the geodesics on the manifold. Subsequently, we will outline our modifications in this work.

For $T$ time steps, $0 = t_0 < t_1 < \ldots < t_{T-1} < t_T = 1$, and discrete time interval $\delta t = (t_{i+1}-t_{i}) = 1/T$, we consider an approximate discretization of the curve $\gamma(t)$ on the latent manifold, $\mathcal{M}$, as $z_0, z_1,\ldots, z_T$, i.e., $z_i = \gamma(t_i)$. The smooth map $g$ now gives a discrete path on the data manifold  as $g(z_0), g(z_1),\ldots, g(z_T)$. Using forward finite differences, we get an approximation to the velocity of the curve $g(\gamma(t))$ at $t_i$ as 
\begin{align*}
    \frac{d g (\gamma(t))}{d t} \mid_{t = t_i} \; \approx \; \frac{g (\gamma(t_{i+1})) - g (\gamma(t_{i}))}{t_{i+1} - t_{i}} = \frac{g(z_{i+1}) - g(z_i)}{\delta t}.
\end{align*}
Now, the discrete analog to the energy functional (\ref{eq:energy}) is given by 
\begin{align*}
    E(\mathbf{z}) = \frac{1}{2} \sum\limits_{t=0}^{T-1} \frac{1}{\delta t} \left\| g(z_{i+1})- g(z_i)\right\|^2, \quad \text{where} \quad \mathbf{z} = \begin{bmatrix}
        z_0 & z_1 & \ldots & z_T 
    \end{bmatrix}.
\end{align*}
Fixing $z_0$ and $z_T$ as the specific start and end points for the geodesic path, we aim to minimize this discrete geodesic energy through gradient descent applied to the intermediate points on the curve, $z_1, ..., z_{T-1}$. The gradient with respect to $z_i$
is thus
\begin{align*}
\frac{\partial}{\partial z_i} E(\mathbf{z}) = \frac{1}{\delta t} J_g ^T (z_i) (g(z_{i+1})- 2g(z_i) + g(z_{i-1})).
\end{align*}

Instead of utilizing $J_g^T$ for calculating gradients, \citet{RiemannianGeometryDeep_2018} opted for the faster-to-compute Jacobian of the encoder, $J_e$, as they considered the former to be computationally expensive. The resulting modified gradient can then be written as
\begin{align}
\label{eq:discrete_eta}
\eta_i = \frac{1}{\delta t} J_e (z_i) (g(z_{i+1})- 2g(z_i) + g(z_{i-1})).
\end{align}

However, to produce more faithful geodesics, we find that using the exact decoder is crucial. To achieve this, instead of forming the Jacobian explicitly, we access it only through Jacobian-vector products. For this, we reorganize the original gradient as
\begin{align}
\label{eq:our_modification}
\frac{\partial}{\partial z_i} E(\mathbf{z}) = \frac{1}{\delta t} \left( \left( g(z_{i+1}) - 2g(z_i) + g(z_{i-1}) \right)^T J_g(z_i) \right)^T
\end{align}
This reorganization enables the computation using vector-Jacobian product, which is more computationally efficient than explicitly computing the Jacobian matrix $J_g$. We use \cref{eq:our_modification} in place of the modified gradient (\ref{eq:discrete_eta}) in  \cref{alg:geodesic}. By doing this, we do not compromise the use of the generator function, which is the core surface model in our work. The resulting modified geodesic path algorithm is given in \cref{alg:geodesic}.
\begin{algorithm}
\caption{Geodesic Path}
\begin{algorithmic}[1]
\label{alg:geodesic}
\INPUT Two points, $z_0, z_T \in \mathcal{Z} $
\STATE $\mathbf{z}^{(0)} = \{z_i^{(0)}\}_{i=0}^{T}$ as linear interpolation between $z_0$ and $z_T$
\smallskip 
\FOR{$k = 0, 1, \ldots$}
\smallskip 
\FOR{$i \in \{1, \ldots, T - 1\}$}
\smallskip 
\STATE Compute the gradient $\displaystyle \frac{\partial}{\partial z_i} E(\mathbf{z}^{(k)})$ using \cref{eq:our_modification}
\smallskip 
\STATE $\displaystyle z_i^{(k+1)} = z_i^{(k)} - \alpha^{(k)}\frac{\partial}{\partial z_i} E(\mathbf{z}^{(k)})$
\smallskip 
\ENDFOR
\ENDFOR
\OUTPUT Discrete geodesic path $z_0, z_1, \ldots, z_T \in \mathcal{Z}$
\end{algorithmic}
\end{algorithm}

For our experiment in \cref{sec:exp}, we employ backtracking line search with Armijo-Goldstein condition \citep{nocedal1999numerical}, to determine the appropriate step size, $\alpha^{(k)}$, for each gradient descent step. The outer loop is terminated after a maximum of 300 iterations or if  $\left| \Delta E^{(k)} - \Delta E^{(k-1)} \right| \leq 0.001 \Delta E^{(k)}$ where  $\Delta E^{(k)} = \sum_i \|\partial E(\mathbf{z}^{(k)})/\partial z_i\|^2$.

\section{Details on the VAE Architecture}
\label{sec:appendix_b}

The VAE used in our experiments, as specified in the code repository, relies intensively on residual blocks and certain activations that allows the decoder to impose a valid Riemannian metric. Table \ref{table:VAE_Arch} shows details of the architectural choices in the VAE model.\\

\begin{table}[h]
\centering
\begin{tabular}{|c|c|c|c|}
\hline
\textbf{Module} & \textbf{Layer} & \textbf{Output Shape} & \textbf{Details} \\
\hline
\multicolumn{4}{|c|}{\textbf{Encoder}} \\
\hline
Input & - & $3 \times 192 \times 192$ & - \\
Conv In & Conv2d & $64 \times 192 \times 192$ & Kernel: 3, Stride: 1, Padding: 1, Activation: SiLU \\
\hline
ResDown1 & Conv2d x2, BN x2 & $128 \times 96 \times 96$ & $\begin{array}{c} \text{Conv1: } \text{Kernel: 3, Stride: 2, Padding: 1} \\ \text{Conv2: } \text{Kernel: 3, Stride: 1, Padding: 1} \\ \text{Activation: SiLU} \end{array}$ \\
\hline
ResDown2 & Conv2d x2, BN x2 & $256 \times 48 \times 48$ & Same as above \\
\hline
ResDown3 & Conv2d x2, BN x2 & $512 \times 24 \times 24$ & Same as above \\
\hline
ResDown4 & Conv2d x2, BN x2 & $512 \times 12 \times 12$ & Same as above \\
\hline
ResBlock & Conv2d x2, BN x2 & $512 \times 12 \times 12$ & Kernel: 3, Stride: 1, Padding: 1, Activation: SiLU \\
\hline
Mu & Conv2d & $64 \times 12 \times 12$ & Kernel: 1, Stride: 1, Padding: 0 \\
Log Var & Conv2d & $64 \times 12 \times 12$ & Kernel: 1, Stride: 1, Padding: 0 \\
\hline
\multicolumn{4}{|c|}{\textbf{Decoder}} \\
\hline
Conv In & Conv2d & $512 \times 12 \times 12$ & Kernel: 1, Stride: 1, Padding: 0, Activation: ELU \\
\hline
ResUp1 & Upsample, Conv2d x2, BN x2 & $512 \times 24 \times 24$ & $\begin{array}{c} \text{Upsample: } \text{Scale: 2} \\ \text{Conv1: } \text{Kernel: 3, Stride: 1, Padding: 1} \\ \text{Conv2: } \text{Kernel: 3, Stride: 1, Padding: 1} \\ \text{Activation: ELU} \end{array}$ \\
\hline
ResUp2 & Upsample, Conv2d x2, BN x2 & $256 \times 48 \times 48$ & Same as above \\
\hline
ResUp3 & Upsample, Conv2d x2, BN x2 & $128 \times 96 \times 96$ & Same as above \\
\hline
ResUp4 & Upsample, Conv2d x2, BN x2 & $64 \times 192 \times 192$ & Same as above \\
\hline
Conv Out & Conv2d & $3 \times 192 \times 192$ & Kernel: 3, Stride: 1, Padding: 1, Activation: Tanh \\
\hline

\end{tabular}
\caption{Detailed architecture of the VAE, showcasing the configurations of layers within the Encoder and Decoder modules.}
\label{table:VAE_Arch}
\end{table}

\section{Limitations of our Approach}
\label{sec:appendix_c}
While our approach bridges the gap between the robustness of attribution maps under adversarial conditions and the perceptual alignment and intuitiveness of explanations, it requires the use of VAEs to capture the underlying data manifold. Two sources of complexity arise from this setting: the training of VAEs and, subsequently, the geodesic computations necessary to generate the path of attribution in MIG. Future directions could aim to capture the Riemannian structure or specify the Riemannian metric needed to compute geodesics in ways that are less costly than using the VAE setup.

\end{document}